# Federated Learning over Human-Body Communication for On-Body Edge Intelligence: A Survey, Taxonomy, and BodyFed-HBC Scheduling Vignette

**Koffka Khan**
**Department of Computing and Information Technology**
**The University of the West Indies, St. Augustine, Trinidad and Tobago**

*Preprint*

**Abstract**

Human-body communication (HBC) is a promising physical substrate for wearable body-area networks because it can localize communication around the body and reduce the burden of conventional radio links. Federated learning (FL) is a promising learning substrate because it can reduce raw-data centralization for physiological and behavioral sensing. Yet these two literatures remain weakly connected: FL for wearables usually abstracts the communication layer, whereas HBC research usually abstracts learning and model-update traffic. This article surveys the intersection of HBC, wireless body-area networks, wearable FL, Internet-of-Bodies privacy, and edge-intelligence optimization. We propose a taxonomy that distinguishes intra-body, body-hub, cross-user, and clinical-cloud FL deployments, and we identify the open problem of body-channel-aware FL: learning protocols whose client selection, update compression, and aggregation are controlled by posture-dependent HBC links, residual energy, sensor memory, and privacy risk. To make the research agenda concrete, we introduce BodyFed-HBC as a reference architecture and provide an optimization formulation and scheduling algorithm. We further specify a reproducible simulation vignette that combines public wearable datasets with empirical body-coupled-communication signal-loss models. The article concludes with open datasets, evaluation metrics, limitations, and research directions for computer scientists working above the hardware layer.

## Introduction

Wearable sensing is moving from episodic monitoring toward continuous on-body edge intelligence. Smart watches, chest patches, earables, textile sensors, ankle gait sensors, and implant-adjacent gateways can capture physiological, inertial, and contextual signals at a rate that is useful for activity recognition, stress detection, rehabilitation monitoring, arrhythmia screening, sleep analysis, and occupational safety. These systems sit at the intersection of the Internet of Things (IoT), wireless body-area networks (WBANs), mobile health, and the emerging Internet of Bodies. They also create a difficult systems question: how should learning be performed when data are personal, continuous, multimodal, and generated by devices with small batteries, small memory, and posture-dependent links?

Human-body communication (HBC), also called intra-body communication or body-coupled communication (BCC) in related communities, uses the human body as part of the signal propagation medium. HBC has been studied for WBANs because it can reduce radiative leakage, electromagnetic interference, and energy expenditure relative to conventional radio links in certain operating regimes (Seyedi et al. 2013; Maity et al. 2019; Yang et al. 2022). Electro-quasistatic HBC (EQS-HBC) and galvanic/capacitive BCC are especially relevant for short-range wearable-to-wearable links. However, HBC is not a perfect channel. Body location, posture, motion, environment, termination impedance, inter-body coupling, and frequency all influence channel loss and reliability (Maity et al. 2019; Nath et al. 2021; Ormanis, Medvedevs, et al. 2023; Yang et al. 2022). A learning protocol that ignores these variations may select clients at the wrong time, spend energy on retransmissions, and bias the model toward sensors with easier links.

Federated learning (FL) provides a complementary abstraction. In FL, clients compute local model updates and an aggregator combines them without requiring raw training data to be centrally collected (McMahan et al. 2017; Kairouz et al. 2021). FL has already been applied to human activity recognition (HAR), wearable healthcare, Internet of Medical Things (IoMT) security, mobile health, and personalized sensing (Sozinov, Vlassov, and Girdzijauskas 2018; Chen et al. 2020; Aouedi et al. 2024; Rani, Kataria, et al. 2023). The motivation is compelling: gait, heart rhythms, electrodermal activity, sleep patterns, and daily routines can be private and identifying. Nevertheless, most wearable FL work treats the communication substrate as generic wireless networking or as a simple uplink bandwidth constraint. That abstraction is reasonable for phone-to-cloud FL, but it is incomplete for intra-WBAN systems in which the model-update path itself is a human body channel.

This article argues that the missing research area is *body-channel-aware federated learning*: FL for on-body edge intelligence in which client selection, update size, compression, aggregation, and participation fairness are co-designed with HBC channel state, posture, energy, and sensor-level hardware constraints. We do not claim that every wearable task should use intra-WBAN FL. In many cases, it is better to stream features to a trusted hub, perform local inference, or train a personal model. FL becomes compelling when (i) raw streaming is continuous and expensive, (ii) physiological streams should not be centralized, (iii) multiple on-body locations have complementary data, (iv) the body hub later participates in cross-user FL, or (v) a model must adapt locally while limiting data movement.

The article is written for computer scientists who work above the analog hardware layer. It therefore does not require the reader to design an HBC transceiver. Instead, it explains what a computer scientist needs from the HBC layer: channel-loss distributions, packet-success models, energy-per-bit estimates, posture labels, and leakage assumptions. The article also introduces BodyFed-HBC as a reference architecture and scheduling vignette that can be implemented in a Jupyter environment using public wearable datasets and empirical BCC signal-loss data. The goal is to create a realistic bridge between survey, research agenda, and reproducible computational exploration.

## Contributions and novelty

The novelty of this survey is not that FL exists for wearables or that HBC exists for WBANs. Both are established. The contribution is the synthesis of these areas into a body-channel-aware learning framework with a reproducible computational path. Specifically, the article provides:

1. a scoping survey of FL for wearables, HBC/BCC, WBAN edge intelligence, and IoT privacy;
2. a taxonomy separating intra-WBAN FL, body-hub FL, cross-user FL, clinical/vendor FL, and hybrid split/distillation variants;
3. a precise answer to “why FL here?” by distinguishing HBC link privacy from raw-data and model-update privacy;
4. the BodyFed-HBC reference architecture, including a constrained optimization formulation and a body-channel-aware scheduling algorithm;
5. a computational vignette blueprint using PAMAP2, MHEALTH, WESAD, and a public BCC signal-loss dataset;
6. a research agenda identifying missing datasets, trace formats, metrics, and collaboration points between computer scientists and HBC hardware researchers.

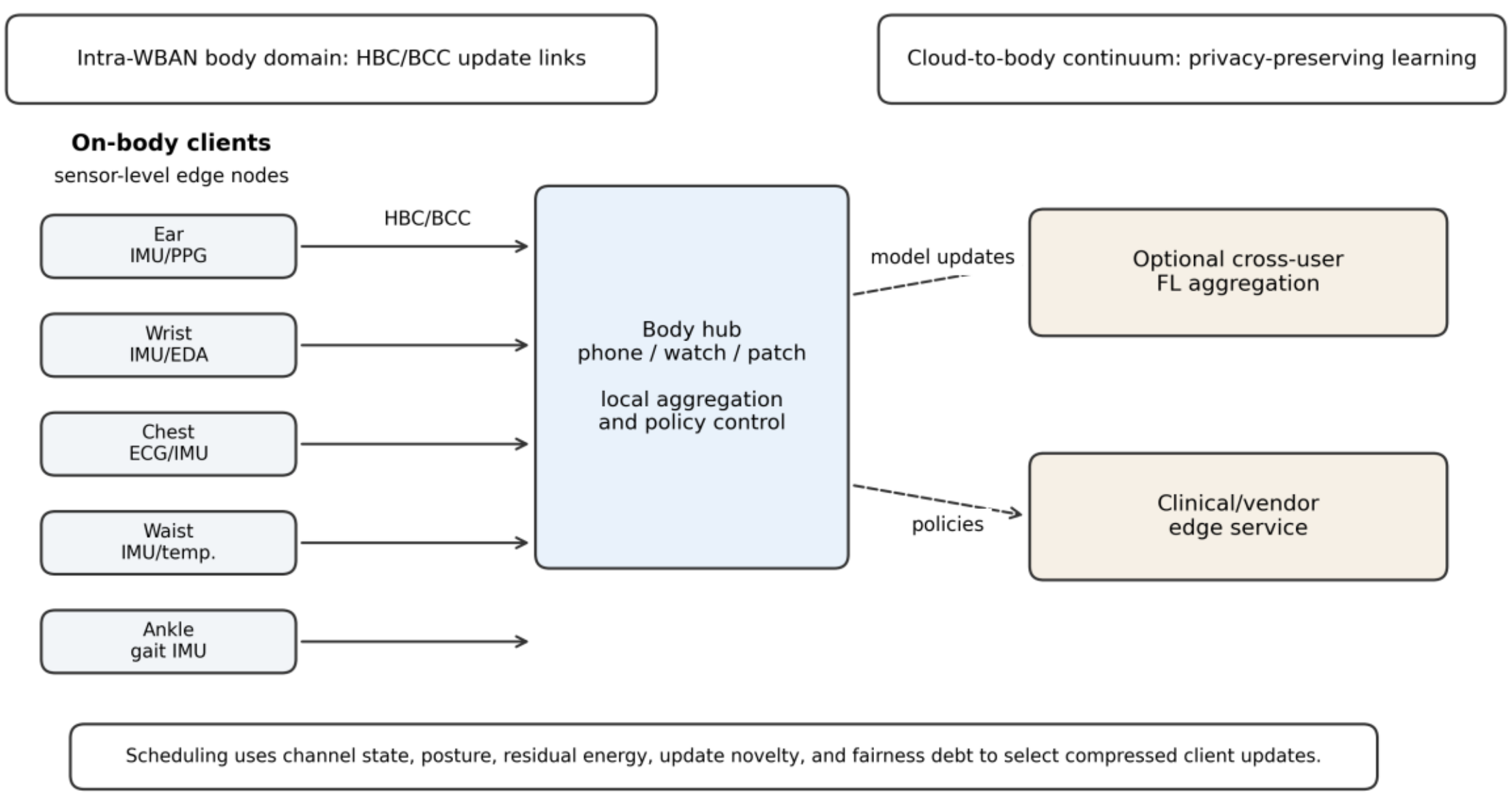

*Figure 1. BodyFed-HBC reference architecture. On-body clients transmit compressed model updates or learning statistics to a body hub through HBC/BCC links. The optional cloud or clinical server participates only in cross-user or cross-institution FL; the intra-WBAN problem already exists inside one body domain.*

## Review protocol and scope

This article is a scoping review and research agenda, not a PRISMA-style systematic medical review. The objective is to identify a technically coherent research gap and translate it into an actionable design framework. The reviewed literature was organized into five families: (i) FL foundations and wireless FL; (ii) FL for wearable HAR, mobile health, and IoMT; (iii) HBC/BCC/EQS-HBC physical-layer and channel-measurement work; (iv) WBAN, Internet-of-Bodies, and edge-intelligence systems; and (v) privacy, security, and reproducibility practices for health-related IoT.

The guiding search strings were combinations of “federated learning” with “human activity recognition”, “wearables”, “Internet of Medical Things”, “wireless body area network”, “edge intelligence”, “client selection”, and “privacy”, together with combinations of “human body communication”, “body-coupled communication”, “intrabody communication”, “electro-quasistatic”, “body area network”, “channel loss”, and “wearable-to-wearable”. Inclusion favored peer-reviewed papers, public datasets, surveys, and works with explicit system models. Exclusion removed work that used the phrase “body area network” only as a generic medical-IoT example without body-channel or wearable details.

The resulting scope is intentionally above the device-circuit layer. Transceiver circuits, electrode geometry, and analog front-end design are treated only to the extent that they determine the abstractions needed by learning and scheduling algorithms: link loss, packet reliability, rate, energy per bit, leakage radius, and posture sensitivity. This is the level at which a computer-science contribution can be credible without claiming hardware novelty.

## Background

### Human-body communication and body-coupled communication

In HBC/BCC, the human body participates in signal propagation. Depending on the modality, the transmitter and receiver may be coupled galvanically through electrodes or capacitively through the body and environment. Seyedi et al. provide an early survey of intra-body communication for body-area network applications (Seyedi et al. 2013). Later work on EQS-HBC developed biophysical channel models and showed how measurement setup, receiver termination, and wearable prototypes affect measured loss (Maity et al. 2019). BodyWire demonstrated wearable-to-wearable EQS-HBC and characterized channel loss under different environment, posture, and body-location conditions (Yang et al. 2022). Inter-body coupling work further showed that a nearby body can become a coupling surface, which complicates simplistic claims that HBC is automatically private or secure (Nath et al. 2021).

For the present survey, four abstractions matter. First, HBC links are location-dependent: wrist-to-chest, ankle-to-waist, and ear-to-phone links may not have the same loss. Second, links are posture-dependent: arm position, sitting, walking, or lying can change the body-channel graph. Third, link failures can be correlated: motion that weakens one distal link may also weaken another. Fourth, HBC can reduce some local radio leakage, but it does not remove privacy risks from model updates, participation metadata, or the cloud-facing part of the system.

### Federated learning for wearables

FL trains a shared model from distributed clients by repeatedly distributing a model, computing local updates, and aggregating those updates (McMahan et al. 2017). In mobile and wearable sensing, FL is attractive because data can remain on the device or personal hub while global or personalized models improve. Early HAR work showed that FL can be applied to smartphone/wearable activity recognition and that model complexity influences communication cost (Sozinov, Vlassov, and Girdzijauskas 2018). FedHealth extended FL to wearable healthcare and personalization (Chen et al. 2020). Recent surveys document substantial FL-HAR work on aggregation, personalization, non-IID data, robustness, concept drift, and edge-device limits (Aouedi et al. 2024; Grataloup and Kurpicz-Briki 2024).

This literature establishes that FL for wearables is no longer a blank area. Therefore, a new paper should not claim novelty merely from combining FL with health data. The underexplored question is more specific: how should FL be scheduled when each on-body client's ability to send an update depends on a body-channel state that changes with posture and movement?

### WBAN edge intelligence and the Internet of Bodies

WBANs typically use body-worn and near-body devices to sense, process, and transmit health or activity information. IoMT and Internet-of-Bodies deployments extend this model to clinical, cloud, and vendor systems. Edge intelligence is valuable because raw streams can be high-rate and private, while decisions such as fall alerts, arrhythmia warnings, or rehabilitation feedback may need low latency. The body hub - for example a smartphone, smartwatch, chest patch, hearing aid, or belt node - is a natural aggregation point. HBC changes the communication layer between sensors and the hub; FL changes what is communicated.

## Literature map and gap analysis

Table 1 summarizes the literature families and their typical assumptions. Figure 2 visualizes the coverage gap. The key observation is that no single mature line of work simultaneously treats wearable FL, posture-dependent HBC, constrained on-body sensor memory, update compression, fairness across body locations, and model-update privacy as a coupled systems problem.

| Literature family | Typical contribution | Typical limitation for this problem | Relevance to BodyFed-HBC |
|---|---|---|---|
| FL foundations | FedAvg, secure aggregation, personalization, robustness, non-IID optimization (McMahan et al. 2017; Kairouz et al. 2021; Li et al. 2020; Bonawitz et al. 2017) | Communication often abstracted as bandwidth, delay, or client availability | Provides the optimization and aggregation base. |
| Wearable FL and HAR | FL for activity, stress, health, and personalization (Sozinov, Vlassov, and Girdzijauskas 2018; Chen et al. 2020; Aouedi et al. 2024) | Usually assumes generic device-to-server connectivity or smartphone clients | Shows why raw body data should be minimized. |
| IoMT/WBAN FL | FL for medical IoT security, monitoring, and resource allocation (Rani, Kataria, et al. 2023; Guo et al. 2020) | Often studies network or cloud-edge tiers rather than HBC as the intra-body update bus | Provides healthcare-IoT motivation and privacy context. |
| HBC/BCC | Channel models, loss measurement, security, transceivers, inter-body coupling (Seyedi et al. 2013; Maity et al. 2019; | Usually does not model FL update | Provides the body- |

| Literature family | Typical contribution | Typical limitation for this problem | Relevance to BODYFED-HBC |
|---|---|---|---|
| physical layer | Nath et al. 2021; Yang et al. 2022; Ormanis, Medvedevs, et al. 2023) | traffic or learning convergence | channel abstraction. |
| Wireless FL | Client selection, over-the-air aggregation, energy-aware FL, resource allocation | Mostly radio-centric and not posture/body-location specific | Provides algorithms that can be specialized to HBC. |
| BODYFED-HBC gap | HBC-aware FL scheduling, compression, aggregation, and reproducible vignettes | Requires measured HBC traces or careful emulation | Target of this article. |

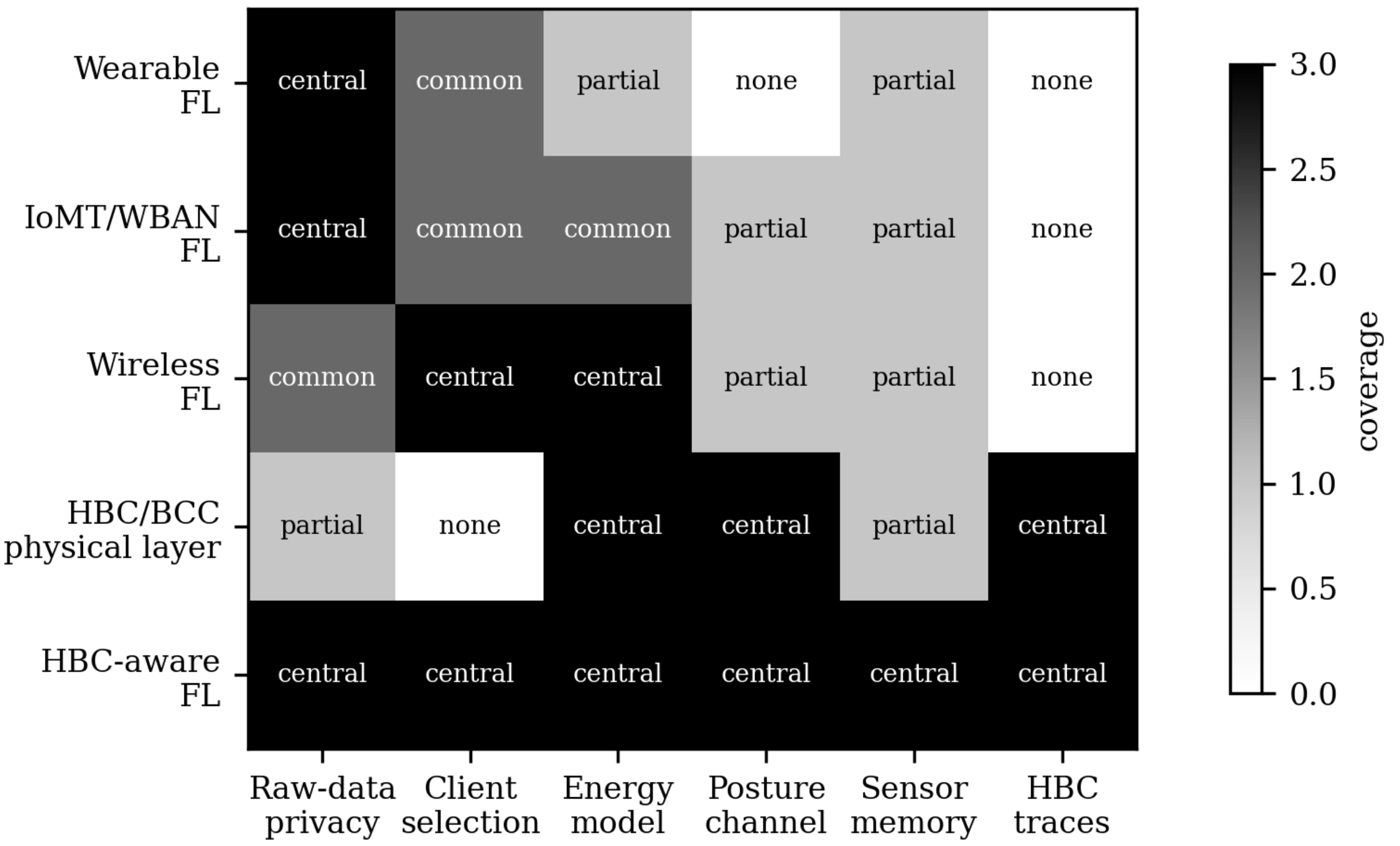

*Figure 2. Coverage map of related literatures. The final row represents the proposed research agenda rather than an established mature literature.*

## Why FL over HBC? A layered privacy and efficiency argument

A common objection is that if HBC already confines communication near the body, FL may be unnecessary. This objection conflates two layers. HBC protects, or at least changes, the local physical communication substrate. FL reduces the need to centralize raw training data. A posture-dependent HBC link can help move updates from a wrist or chest patch to a body hub, but it does not protect the data after the hub uploads raw streams to a cloud service. Conversely, FL can reduce raw-data upload, but it does not guarantee that local HBC transmissions are reliable, energy-efficient, or immune to eavesdropping and coupling.

The strongest privacy assets are not packet payloads between two trusted sensors on the same user. The privacy assets are longitudinal body-generated data: ECG and PPG waveforms, gait, tremor, stress responses, respiration, sleep rhythms, activity routines, and context. These streams can reveal health status, identity, disability, medication effects, workplace behavior, and daily routines. FL is useful when many users, clinics, or body hubs need to improve a model without pooling these streams centrally. Within one body, FL or split/distillation variants can also reduce continuous raw streaming from sensor nodes to the hub, but this benefit must be compared against the energy cost of local training and update transmission.

Figure 3 summarizes the layered model. HBC can reduce some local radio leakage; FL can reduce raw-data centralization; secure aggregation and differential privacy can reduce model-update leakage; robust aggregation can reduce poisoning. None of these mechanisms alone is sufficient.

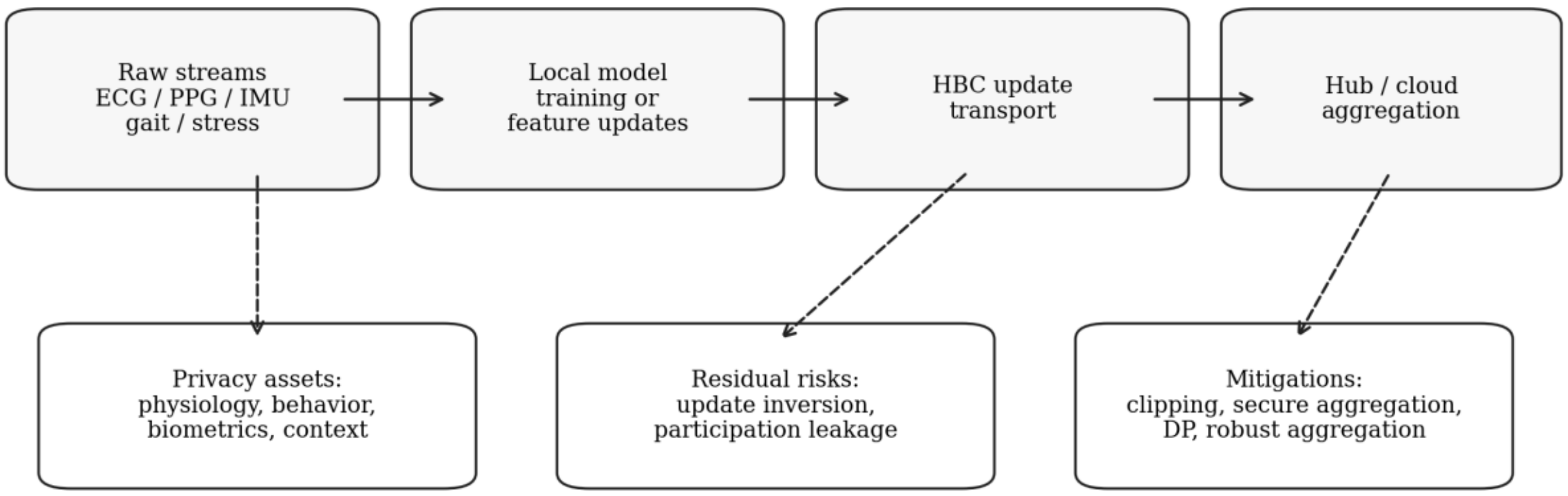

*Figure 3. Layered privacy model. HBC and FL address different risks. HBC changes the local channel; FL reduces raw-data centralization; privacy-preserving FL mechanisms address residual model-update leakage.*

## Taxonomy of learning architectures

Table 2 distinguishes the main learning architectures. This distinction is important because not all of them justify the same novelty claim or require the same experiments.

| Architecture | Clients and aggregator | When it is useful | Main challenge |
|---|---|---|---|
| Local-only TinyML | Each sensor or hub trains/infers locally; no federation | Personal models, no population learning, trusted hub | Limited data diversity and drift adaptation. |
| Raw streaming to hub | Sensors stream raw or feature data to a body hub | Short sessions, simple trusted systems, no local training | Continuous communication cost and raw-data exposure at hub. |
| Intra-WBAN FL | On-body sensors are clients; body hub aggregates | Multi-location sensing with constrained links; raw-stream reduction | Sensor memory, local labels, update size, HBC reliability. |
| Split learning / feature FL | Sensors compute features or early layers; hub completes learning | Sensors too small for full FL | Partition choice, activation leakage, latency. |
| Body-hub FL | Each user hub is one FL client; server aggregates across users | Population learning without raw cloud upload | Non-IID users, privacy leakage, client availability. |
| Hierarchical FL | Sensors to body hub, then hubs to server | Combines intra-body and cross-user learning | Multi-level scheduling and privacy accounting. |
| Federated distillation | Clients send logits, prototypes, or teacher outputs | Heterogeneous sensors/models | Public/reference data and distillation leakage. |

For a computer scientist without HBC hardware, the most credible first contribution is a survey plus a BodyFed-style simulation vignette. A full experimental systems paper would require measured HBC traces, packet-level reliability, and energy profiles. A review paper can instead define the architecture, specify what traces are needed, and demonstrate the scheduling logic with public data.

## From cross-domain physics to traditional optimization principles

Four recent Nature-family papers outside WBAN and FL are not direct prior art for HBC-aware learning. Citing them as if they directly validate BodyFed-HBC would be inappropriate. They can, however, be used as design analogies that translate into standard optimization principles. Table 3 gives a conservative mapping.

| Cross-domain result | Conservative interpretation | Optimization principle | Use in BodyFed-HBC |
|---|---|---|---|
| Average topological phase in disordered Rydberg arrays (Yue et al. 2026) | Useful structure can persist in an ensemble even when individual realizations are disordered. | Distributionally robust optimization. | Optimize expected learning utility over posture/channel disorder, not only the current link snapshot. |
| Standing-wave visibility of topology in a spinning fluid (Singh et al. 2026) | Global structure can be inferred from wave-response signatures. | Graph observability and topology-aware scheduling. | Use HBC pilots to estimate a body-channel graph and schedule by graph coverage. |
| Ion correlations in diffusion-limited synthesis (Karan et al. 2026) | Pairwise local reasoning misses correlated transport and kinetic bottlenecks. | Covariance-aware resource allocation. | Penalize correlated link failures and redundant updates; reward update diversity. |
| De novo chemo-optogenetics with rational switch/binder design (Miyazaki et al. 2026) | Function can be programmed by jointly designing a switch and a selective binding partner. | Switched-hybrid control. | Treat client participation as a gated decision triggered by posture, energy, novelty, and privacy state. |

These principles are deliberately traditional. The point is not to import quantum, fluid, materials, or chemical mechanisms into WBANs. The point is to discipline the algorithm design using well-known tools such as robust optimization and constrained resource allocation (Ben-Tal, El Ghaoui, and Nemirovski 2009; Boyd and Vandenberghe 2004): robust learning for disorder, graph scheduling for channel topology, covariance-aware selection for correlated failures, and switched control for context-dependent participation.

# The BodyFed-HBC reference architecture

## System model

Consider an intra-WBAN with on-body clients $i \in \mathcal{N} = \{1, \dots, N\}$ and a body hub. Client $i$ owns local data $\mathcal{D}_i^t$ generated by a sensor location and modality, for example wrist IMU, chest ECG, ankle IMU, or ear PPG. The hub maintains a model $w^t \in \mathbb{R}^d$ at round $t$. The local empirical risk is

$$F_i(w) = \frac{1}{n_i} \sum_{(x,y) \in \mathcal{D}_i} \ell\,(w; x, y),$$

and the body-level objective is

$$F(w) = \sum_{i=1}^{N} p_i\, F_i(w), \qquad p_i = \frac{n_i}{\sum_j n_j}.$$

Client $i$ computes a local update $\Delta w_i^t$ or a lower-dimensional statistic such as a prototype, feature-head update, or distillation vector.

Let $\xi^t$ denote body context, including posture and motion. The HBC channel from client $i$ to the hub has path loss $L_i^t = L_i(\xi^t)$, packet error rate $\epsilon_i^t$, rate $R_i^t$, expected retransmission factor $\rho_i^t = 1/(1 - \epsilon_i^t)$ when $\epsilon_i^t < 1$, and energy-per-useful-bit $\eta_{i,\text{bit}}^t$. The total round energy is

$$\hat{E}_i^t = E_{i,\text{train}}^t + E_{i,\text{tx}}^t + E_{i,\text{rx}}^t.$$

If $s_i^t$ parameters or statistics are sent with $q_i^t$ bits per value, then

$$E_{i,\text{tx}}^t = \eta_{i,\text{bit}}^t s_i^t q_i^t \rho_i^t.$$

The node must satisfy

$$\hat{E}_i^t \le B_i^t, \qquad M_i(w^t, q_i^t) \le M_i^{\max},$$

where $B_i^t$ is the current energy budget and $M_i^{\max}$ is the memory bound.

## Optimization objective

At each round, the hub chooses binary participation variables $z_i^t \in \{0,1\}$, quantization or sparsification levels $q_i^t$, and aggregation weights $a_i^t$. A high-level constrained objective is

$$\begin{aligned}\min_{\{z,q,a\}} \quad & \mathbb{E}_{\xi\sim\mathcal{P}}[F(w^T;\xi)] + \lambda_E \sum_{t,i} z_i^t \hat{E}_i^t + \lambda_T \sum_{t,i} z_i^t \hat{T}_i^t \\ & + \lambda_U \sum_t \mathcal{U}(z^t) + \lambda_P \sum_{t,i} z_i^t \hat{P}_i^t,\end{aligned}$$

subject to

$$\begin{aligned}& \sum_i z_i^t \le K, \quad \hat{E}_i^t \le B_i^t, \quad \hat{T}_i^t \le T_{\max}, \\ & \epsilon_i^t \le \epsilon_{\max}, \quad M_i(w^t, q_i^t) \le M_i^{\max}, \\ & \sum_i a_i^t = 1, \quad a_i^t \ge 0, \quad a_i^t = 0 \text{ if } z_i^t = 0.\end{aligned}$$

Here $\mathcal{U}(z^t)$ penalizes unfair exclusion across body locations and $\hat{P}_i^t$ is a privacy-risk proxy, for example update norm after clipping, task sensitivity, or participation metadata risk.

For online scheduling, BodyFed-HBC uses a decomposed utility. Let $v_i^t$ be update value, $d_i^t$ data novelty, $h_i^t$ fairness debt, $c_i^t$ channel cost, $e_i^t$ energy cost, and $p_i^t$ privacy-risk proxy. Let $\Sigma_\Delta^t$ be the covariance of recent updates and $\Sigma_c^t$ the covariance of link failures. The round-wise selection problem is

$$\begin{aligned}\max_{z^t \in \{0,1\}^N} \quad & \sum_i z_i^t \left(\alpha_v v_i^t + \alpha_d d_i^t + \alpha_h h_i^t - \lambda_c c_i^t - \lambda_e e_i^t - \lambda_p p_i^t\right) \\ & + \rho_1 \operatorname{logdet}(I + Z^t \Sigma_\Delta^t Z^t) - \rho_2 z^{t\top} \Sigma_c^t z^t,\end{aligned}$$

where $Z^t = \operatorname{diag}(z^t)$. The log-determinant term rewards update diversity; the covariance penalty discourages selecting clients whose HBC links fail together.

## Algorithm

Algorithm 1 gives the practical greedy version. It is intentionally simple enough to implement in a Jupyter vignette. A later systems paper could replace it with a knapsack solver, a submodular approximation method, contextual bandit, model-predictive controller, or constrained reinforcement learner (Nemhauser, Wolsey, and Fisher 1978).

Send pilot request and estimate  for each client Estimate data value , novelty , fairness debt , energy cost , privacy proxy  Greedily add feasible clients to  by decreasing marginal gain in the round-selection objective until  or no feasible client remains Selected clients train locally and transmit  over HBC Compute reliability scores  Set aggregation weights  for delivered updates Update  Update fairness debt  and selection probability estimate

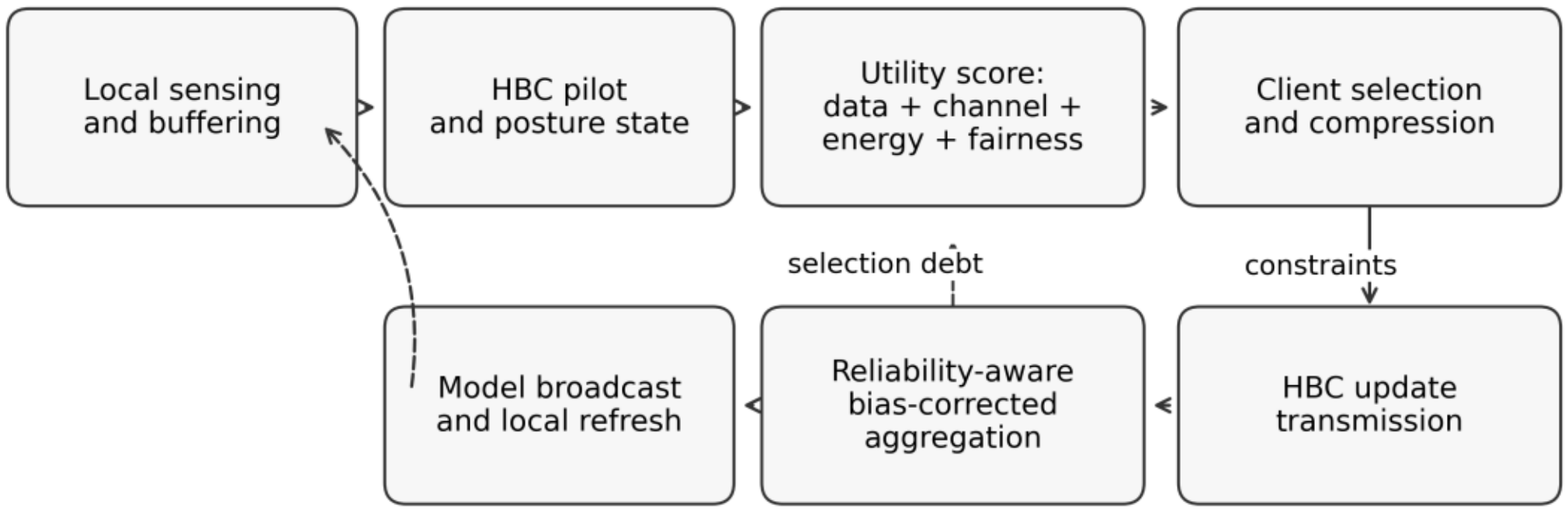

*Figure 4. One BodyFed-HBC round. The learning decision uses sensing context, HBC pilots, energy state, update novelty, and fairness debt before choosing clients and update compression.*

## Bias-corrected aggregation

A purely channel-aware scheduler may repeatedly choose chest or waist nodes because their links are reliable. That can reduce energy but harm activity classes that require distal sensors, such as ankle gait or wrist gestures. BODYFED-HBC uses inverse-propensity reliability-aware aggregation:

$$a_i^t = \frac{n_i^t r_i^t / \max(\pi_i^t, \epsilon_\pi)}{\sum_{j \in \mathcal{S}_t} n_j^t \, r_j^t / \max\left(\pi_j^t, \epsilon_\pi\right)},$$

where $\pi_i^t$ is the moving-average probability of selecting client $i$. This does not remove all bias, but it forces the scheduler to account for the statistical consequences of repeatedly excluding difficult links.

# Reproducible simulation vignette

A review paper becomes more useful if it includes a small, transparent computational vignette. The vignette should not be presented as a hardware validation. Its purpose is to show how public wearable datasets and empirical BCC signal-loss models can be combined to test a body-channel-aware scheduling idea.

## Datasets

Table 4 lists suitable datasets. PAMAP2 is the cleanest primary dataset because it contains wrist/hand, chest, and ankle IMUs plus heart rate for 18 activities across 9 subjects (Reiss and Stricker 2012; UCI Machine Learning Repository 2012). MHEALTH is a smaller and more sandbox-friendly benchmark with sensors at chest, right wrist, and left ankle for 12 activities (Banos et al. 2014; UCI Machine Learning Repository 2014). WESAD is useful for privacy and physiological stress discussion because it contains wrist and chest physiological/motion data from 15 subjects, but the modalities and sampling rates are heterogeneous (Schmidt et al. 2018; UCI Machine Learning Repository 2018). For HBC/BCC, Ormanis et al. provide a public galvanic-coupled BCC signal-loss and bioimpedance dataset recorded from 30 volunteers across 50 kHz to 20 MHz (Ormanis, Medvedevs, et al. 2023). That dataset does not directly provide FL packet traces, but it can support empirical channel-loss distributions.

| Dataset | Role | Useful properties | Limitation | Recommended use |
|---|---|---|---|---|
| PAMAP2 (Reiss and Stricker 2012; UCI Machine Learning Repository 2012) | Main HAR dataset | 18 activities, 9 subjects, wrist/hand, chest, ankle IMUs, heart rate | Larger preprocessing burden; missing values | Main paper benchmark after pipeline is stable. |
| MHEALTH (Banos et al. 2014; UCI Machine Learning Repository 2014) | Sandbox HAR dataset | 10 volunteers, 12 activities, chest, wrist, ankle sensors, 50 Hz | Smaller subject count | First Jupyter implementation and sanity checks. |
| WESAD (Schmidt et al. 2018; UCI Machine Learning Repository 2018) | Physiological privacy dataset | Wrist and chest devices, stress/affect labels, ECG, EDA, respiration, BVP, temperature, acceleration | Heterogeneous modalities and sampling rates | Secondary discussion or distillation/split-learning case. |
| BCC signal-loss dataset (Ormanis, Medvedevs, et al. 2023) | Channel model dataset | 30 volunteers, signal loss and bioimpedance, 50 kHz to 20 MHz | No HAR labels, posture-aligned packet traces, or FL traffic | Link-loss distribution and empirical HBC/BCC emulator. |

Proposed reproducible simulation vignette for a review-plus-agenda article

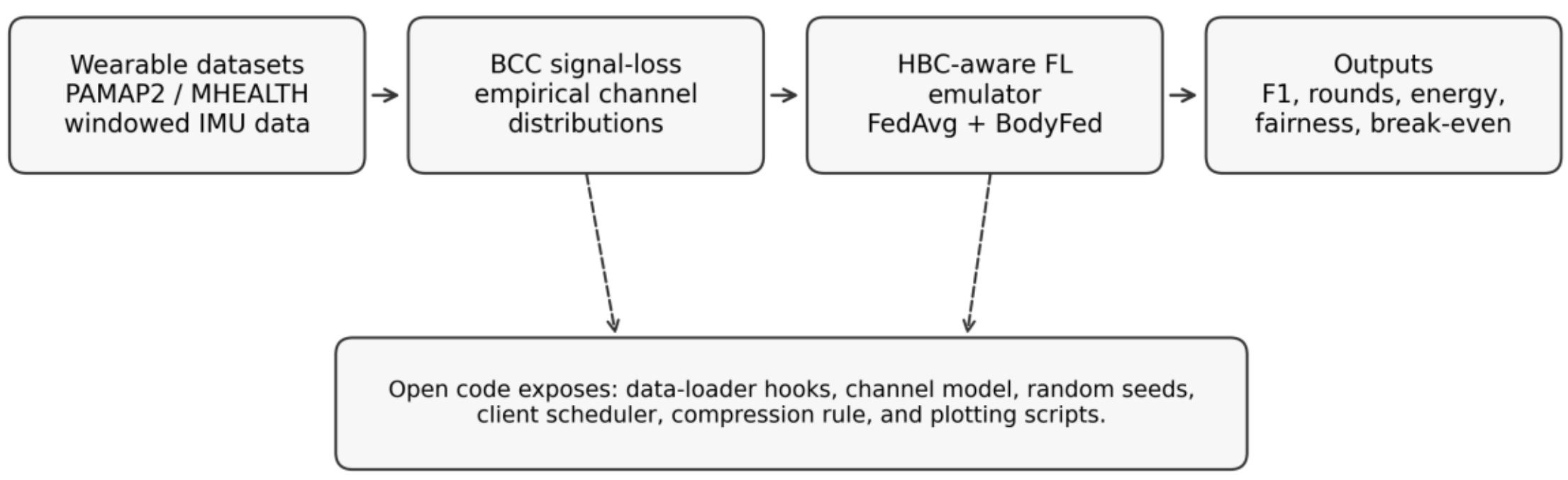

*Figure 5. Computational vignette pipeline. Public wearable datasets provide the learning task; BCC signal-loss data provide empirical body-channel distributions; the scheduler emulates HBC-constrained FL rounds.*

## Vignette protocol

A minimum reproducible protocol is as follows.

1. **Window wearable data.** Segment wrist, chest, and ankle streams into 1–5 second windows with fixed overlap. Normalize per subject or per training split. Use subject-disjoint evaluation where possible.

2. **Define clients.** Treat each body location as an intra-WBAN client: wrist/hand, chest, ankle, and optional heart-rate or physiological clients.

3. **Train tiny models.** Use logistic regression, a small multilayer perceptron, or a tiny 1D convolutional model. For strict FedAvg, use comparable input dimensions across clients. For heterogeneous WESAD-like modalities, use distillation or feature-level FL.

4. **Create an HBC/BCC emulator.** Fit channel-loss distributions from BCC signal-loss data. Map loss to packet success, retransmissions, rate, and energy-per-bit through an explicit link budget. Add posture states using distributions reported in HBC studies or synthetic perturbations clearly labeled as assumptions.

5. **Compare baselines.** Evaluate local-only, centralized upper bound, ideal FedAvg, naive FedAvg over HBC, random client selection, energy-only selection, channel-only selection, data-only selection, and BodyFed-HBC.

6. **Report systems metrics.** Report macro-F1, rounds-to-target-F1, update success rate, cumulative normalized update energy, energy-to-target-F1, worst-location F1, and raw-streaming versus FL break-even.

## Illustrative emulator output

Figure 6 shows deterministic output from a bounded toy emulator included in the source package. It is not a measured hardware result and not a final benchmark. It is included to demonstrate what a reproducible vignette figure should look like. The desirable result is not merely higher accuracy; it is lower energy-to-target-accuracy without excluding difficult body locations.

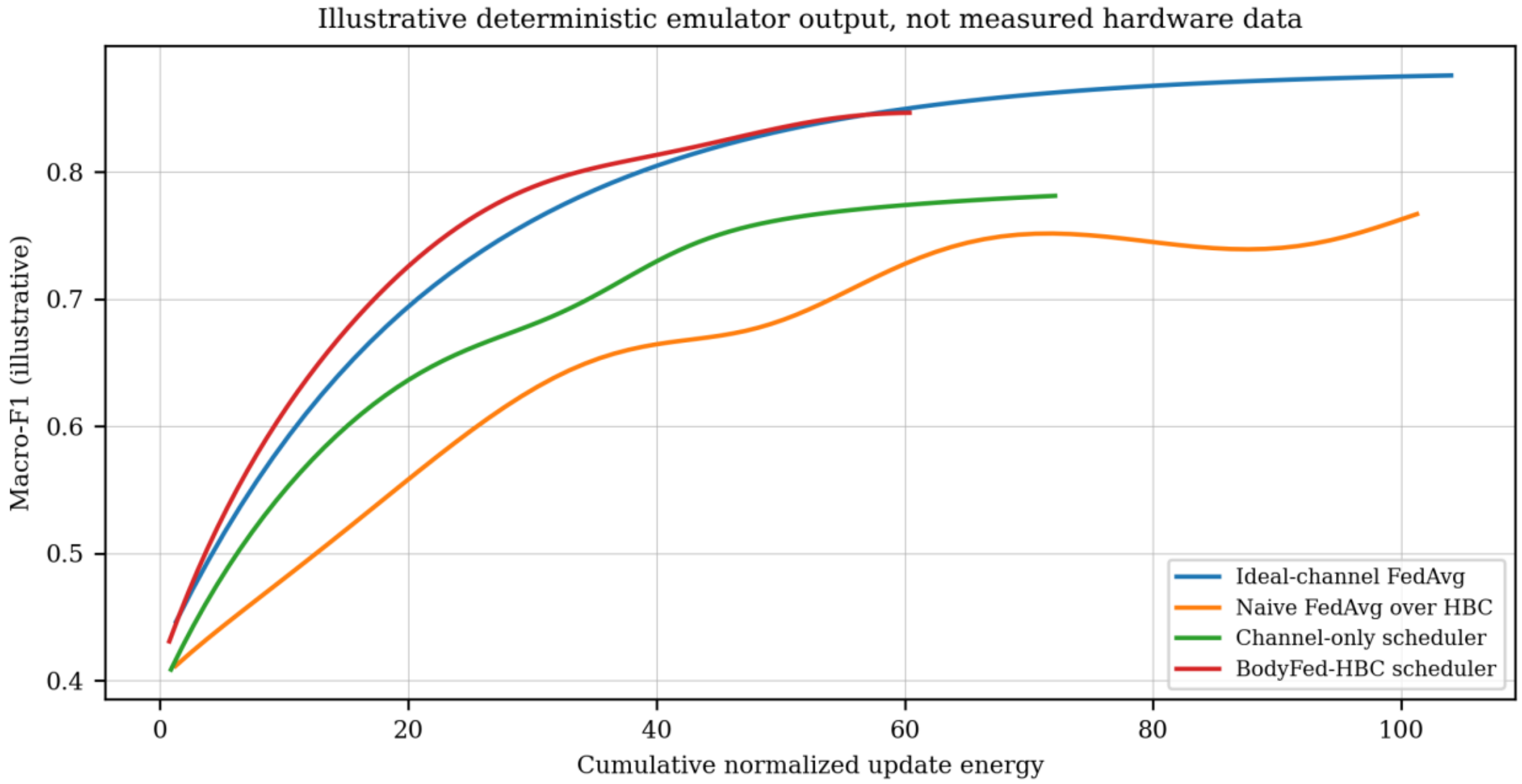

*Figure 6. Illustrative deterministic emulator output. The curves are not measured HBC data and should be replaced by dataset-driven training and measured or public BCC trace replay for submission as an experimental paper.*

## Raw streaming break-even

A reviewer will reasonably ask whether FL is better than simply streaming raw windows to the hub. The vignette should therefore include a break-even calculation. If raw streaming uses sampling rate $f_s$, feature dimension $d_x$, bits per sample $b_x$, duration $T$, and HBC energy per useful bit $\eta_{\text{bit}}$, then

$$E_{\text{stream}} = T f_s d_x b_x \eta_{\text{bit}}.$$

If FL uses $R$ rounds, update size $s$, quantization $q$, retransmission factor $\rho$, and local training energy $E_{\text{train}}$, then

$$E_{\text{FL}} = Rsq\eta_{\text{bit}}\rho + \sum_{r=1}^{R} E_{\text{train}}^{(r)}.$$

FL is attractive only when $E_{\text{FL}}$ is lower than streaming for the task horizon, or when raw-data minimization and cross-user learning justify extra local training cost. This caveat should be stated explicitly.

## Evaluation checklist for future experimental work

Table 5 gives baselines and metrics that would be expected if the review evolves into a full systems paper. The most important baseline is raw streaming to a hub, because it tests the practical value of FL in a one-body domain. The most important fairness metric is worst-location performance, because channel-aware scheduling can otherwise overfit to easy links.

| Category | Baselines or metrics | Purpose |
|---|---|---|
| Learning baselines | Local-only, centralized upper bound, ideal FedAvg, FedProx, personalized FL | Separate learning value from communication effects. |
| Communication baselines | Raw streaming, naive FedAvg over HBC, BLE-like abstract link, no-retransmission link | Test whether HBC-aware FL is better than simpler data movement. |
| Scheduling baselines | Random, round-robin, channel-only, energy-only, data-only, age/fairness-only | Show that multi-factor scheduling is necessary. |
| Compression baselines | Full precision, 8-bit, 4-bit, sign, top-$k$, prototype/logit update | Determine whether model updates fit HBC rate and energy budgets. |
| Learning metrics | Macro-F1, balanced accuracy, subject-disjoint performance, drift recovery | Avoid accuracy artifacts from class imbalance and user leakage. |
| Systems metrics | Energy-to-target-F1, latency, packet success, retransmissions, memory, update size | Make IoT constraints visible. |
| Fairness/privacy metrics | Worst-location F1, selection disparity, update norm leakage proxy, participation leakage | Avoid easy-link bias and overclaiming privacy. |

## Open challenges and research agenda

### Public HBC trace scarcity

The largest barrier is not lack of wearable datasets. It is lack of public, packet-level, posture-labeled HBC traces aligned with wearable tasks. A useful trace format would include body locations, electrode or coupling configuration, carrier/modality, channel loss, packet-success rate, retransmissions, energy-per-bit, posture/activity labels, environment, subject metadata at a privacy-preserving granularity, and calibration details. Without such traces, computer scientists must use empirical loss distributions or synthetic perturbations and clearly label the limitation.

### Client definition in intra-WBAN FL

In normal FL, a client is often a user, phone, hospital, or data silo. In intra-WBAN FL, a client may be a body location, a sensor, a modality, a feature extractor, or a body hub. These choices affect privacy and learning meaning. Treating wrist and ankle sensors as separate clients can be useful for scheduling, but it does not create strong privacy between trusted devices on one person. Treating body hubs as clients is more compelling for cross-user privacy.

### Labels and personalization

Wearable tasks often lack dense labels in real deployment. A credible BodyFed-HBC system may need self-supervised learning, semi-supervised FL, active learning, or personalization. For HBC scheduling, labels are not the only data value. Novelty, drift, uncertainty, rare activities, or change points can trigger updates.

### Model-update leakage

FL does not make data anonymous. Gradients and weights can leak training information (Zhu, Liu, and Han 2019; Geiping et al. 2020). Therefore, a BodyFed-HBC paper should avoid simplistic privacy claims. It should specify whether the threat is local eavesdropping, malicious hub, honest-but-curious cloud server, compromised client, or outside observer. It should also discuss secure aggregation, differential privacy, update clipping, robust aggregation, and poisoning defenses.

### TinyML and split alternatives

Many on-body sensors cannot train full neural networks. BodyFed-HBC should therefore consider alternatives: classifier-head FL, feature-level FL, split learning, federated distillation, prototype sharing, event-triggered updates, and adapter-only tuning. For ultra-low-power nodes, sending a compact statistic may be more realistic than sending dense gradients.

### Ethics and human-subject reporting

On-body systems involve human data. Future work should report how sex/gender, age, health status, and body morphology may affect channel and learning generalizability where relevant and ethically collected. The purpose is not to infer sensitive attributes, but to avoid systems that work only for narrow populations.

## Practical advice for computer scientists

A computer scientist can make a credible contribution without building electrodes if the claims are scoped correctly. A strong review-plus-vignette paper should say: "We provide a taxonomy, optimization formulation, reference scheduler, and reproducible emulation pipeline." It should not say: "We validate an HBC hardware system" unless measurements were collected. The practical path is:

1. start with MHEALTH in Jupyter to debug windowing and sensor-location clients;
2. move to PAMAP2 for the main HAR benchmark;
3. use the BCC signal-loss dataset to fit empirical link-loss distributions;
4. emulate packet success, retransmissions, and update energy;
5. compare BodyFed-HBC with raw streaming and simple scheduling baselines;
6. release code and trace-generation assumptions as supplementary material;
7. collaborate with HBC hardware researchers for a later measurement paper.

This path aligns well with a survey, tutorial, case-study, or open-software/data manuscript. It is less risky than claiming a full experimental HBC-FL systems paper before hardware traces are available.

## Limitations

This article has three limitations. First, the proposed BODYFED-HBC architecture is a reference design and computational vignette, not a measured HBC prototype. Second, public BCC signal-loss data can inform channel distributions but do not replace posture-labeled packet traces collected during wearable tasks. Third, intra-WBAN FL may not always outperform raw streaming or local-only learning. Its value depends on task horizon, update size, compression, local training cost, privacy requirements, and whether the body hub later participates in cross-user FL.

These limitations are also opportunities. The field needs benchmark traces, open emulators, and cross-layer datasets that allow computer scientists, communications researchers, and wearable-system engineers to evaluate the same algorithms.

## Conclusion

Federated learning over human-body communication is not a mature subfield, but it is a coherent research frontier. The broad idea of FL for wearables is already well studied; the underexplored part is body-channel-aware learning in which posture-dependent HBC/BCC links, sensor energy, update size, and body-location fairness shape the FL loop. This article surveyed the relevant literature, separated privacy layers, proposed a taxonomy, introduced

BODYFED-HBC as a reference architecture, formulated the optimization problem, and specified a reproducible simulation vignette that can be executed above the hardware layer. The most defensible immediate manuscript is a survey and research agenda with a transparent computational case study. A full experimental systems paper should follow only after measured HBC traces or a hardware collaborator are available.

## CRediT author statement

Koffka Khan: Conceptualization, Methodology, Formal analysis, Software, Visualization, Writing - original draft, Writing - review and editing.

## Declaration of competing interest

The author declares no known competing financial interests or personal relationships that could have appeared to influence the work reported in this paper.

## Data and code availability

This draft does not report new human-subject measurements. The proposed vignette uses public wearable datasets and public BCC signal-loss data described in Table 4. A future implementation should release the data-loader hooks, fitted channel model, random seeds, client scheduler, compression rules, and plotting scripts in a public repository with a persistent identifier.

## Acknowledgements

No measured hardware results are claimed in this draft. Funding sources, ethics approvals, and institutional acknowledgements should be added if applicable before formal submission.